\documentclass[
]{ceurart}

\usepackage{listings}
\usepackage{graphicx}
\usepackage{booktabs}
\usepackage{multirow}
\usepackage{subcaption}

\begin{document}

\copyrightyear{2025}
\copyrightclause{Copyright for this paper by its authors.
  Use permitted under Creative Commons License Attribution 4.0
  International (CC BY 4.0).}

\conference{Late-breaking work, Demos and Doctoral Consortium, colocated with The 3rd World Conference on eXplainable Artificial
Intelligence: July 09–11, 2025, Istanbul, Turkey}

\title{SHAP-Guided Regularization in Machine Learning Models}

\tnotemark[1]
\tnotetext[1]{This research has partly been funded by the Federal Ministry of Education and Research of Germany and the state of North-Rhine-Westphalia as
part of the Lamarr-Institute for Machine Learning and Artificial Intelligence.}
\tnotemark[1]
\tnotetext[1]{Accepted in Late-breaking works as part of the 3rd World Conference on eXplainable Artificial Intelligence(XAI 2025), proceedings forthcoming.}

\author{Amal Saadallah}[%
orcid=0000-0003-2976-7574,
email=amal.saadallah@cs.tu-dortmund.de,
]

\address{Lamarr Institute for Machine Learning and AI, Dortmund, Germany}

\cortext[1]{Corresponding author.}




\begin{abstract}
  Feature attribution methods such as SHapley Additive exPlanations (SHAP) have become instrumental in understanding machine learning models, but their role in guiding model optimization remains underexplored. In this paper, we propose a SHAP-guided regularization framework that incorporates feature importance constraints into model training to enhance both predictive performance and interpretability. Our approach applies entropy-based penalties to encourage sparse, concentrated feature attributions while promoting stability across samples. The framework is applicable to both regression and classification tasks. Our first exploration started with investigating a tree-based model regularization using TreeSHAP. Through extensive experiments on benchmark regression and classification datasets, we demonstrate that our method improves generalization performance while ensuring robust and interpretable feature attributions. The proposed technique offers a novel, explainability-driven regularization approach, making machine learning models both more accurate and more reliable.

\end{abstract}


\maketitle

\section{Introduction}

As machine learning models become increasingly complex, their interpretability and robustness are critical concerns across various domains, from finance and healthcare to autonomous systems \cite{das2020opportunities}. While deep learning and gradient-boosted trees have shown remarkable predictive power, their black-box nature makes them difficult to trust in high-stakes applications. To address this, explainability techniques such as SHapley Additive exPlanations (SHAP) \cite{molnar2020interpretable} have been widely adopted to quantify feature importance, offering insights into model decisions. However, while SHAP values help interpret trained models \cite{belle2021principles,li2022extracting}, they are rarely incorporated directly into the training process to improve model behavior.

In this work, we introduce SHAP-guided regularization, a novel approach that integrates feature importance constraints into model optimization. Our method introduces two key regularization terms. The first term consists of SHAP entropy penalty – Encourages the model to rely on a sparse, well-distributed subset of important features. The second term is SHAP stability penalty– Ensures that feature attributions remain stable across different samples, reducing sensitivity to small perturbations in the data.
By embedding these explainability-driven constraints into the learning objective, our method enhances both predictive accuracy and interpretability. The framework is applicable to both regression and classification tasks, and first experiments have shown that it is particularly effective for tree-based models such as LightGBM, XGBoost, and CatBoost.

We evaluate our approach on a diverse set of benchmark datasets, comparing its performance against standard models. Our results show that SHAP-guided regularization improves generalization by reducing overfitting to spurious correlations, enhances interpretability by concentrating feature importance on the most relevant predictors, and increases robustness by ensuring stable attributions across samples.

The rest of this paper is structured as follows: Section 2 discusses related works, including SHAP-based model interpretation and feature importance-driven regularization. Section 3 details our SHAP-guided regularization framework and training procedure. Section 4 presents empirical results, demonstrating the effectiveness of our approach across regression and classification tasks. Finally, Section 5 concludes with insights and future directions.

\section{Related Works}
\subsection{Feature Importance and Explainability in Machine Learning}
Interpretability in machine learning has gained significant attention, particularly in domains where model decisions impact critical outcomes, such as finance, healthcare, and autonomous systems. Traditional feature importance measures, such as permutation importance \cite{altmann2010permutation} and Gini importance in decision trees \cite{menze2009comparison}, provide insights into model behavior but often suffer from instability and bias toward correlated features.

SHapley Additive exPlanations (SHAP) \cite{antwarg2021explaining} are a widely used approach that attributes feature importance based on cooperative game theory principles. Unlike other methods, SHAP ensures fair and consistent feature attribution, making it a popular tool for understanding model predictions. However, most applications of SHAP focus on post hoc analysis—explaining trained models—rather than integrating feature attributions into the learning process \cite{molnar2020interpretable}.

\subsection{Regularization for Improved Generalization and Interpretability}
Regularization techniques such as L1 (Lasso) \cite{schmidt2009optimization} and L2 (Ridge) penalties \cite{cortes2012l2} are commonly employed to improve model generalization by controlling feature weights. While these methods help prevent overfitting, they do not explicitly guide the model to focus on the most meaningful features. Other forms of feature selection, such as tree-based pruning \cite{chen2011rough} and attention mechanisms in deep learning \cite{brauwers2021general}, aim to refine model decision-making but often rely on heuristic approaches rather than interpretable attributions like SHAP values.

Some studies have explored feature importance-driven regularization. For instance, Alvarez-Melis and Jaakkola \cite{alvarez2018towards} propose stability-driven constraints to ensure consistent model explanations across similar samples. Meanwhile, Lundberg et al. \cite{lundberg2017unified} discuss the use of SHAP for feature selection but do not incorporate it into the training objective. To our knowledge, no prior work has introduced a SHAP-guided regularization framework that is applicable to both regression and classification tasks while explicitly optimizing for interpretability, stability, and predictive performance.

\subsection{SHAP-Guided Learning: Bridging Interpretability and Optimization}
 A few recent works have begun exploring SHAP-integrated learning.
In \cite{wang2021shapley}, a neural network architecture that incorporates Shapley values as latent representations. This design allows for intrinsic, layer-wise explanations during the model's forward pass, facilitating explanation regularization during training and enabling rapid computation of explanations at inference time. The authors in \cite{sevillano2024x} propose X-SHIELD, a regularization technique that enhances model explainability by selectively masking input features based on explanations. Seamlessly integrated into the objective function, X-SHIELD improves both the performance and interpretability of AI models. SHAPNN \cite{cheng2023shapnn} is a deep learning architecture tailored for tabular data, integrating Shapley values as a regularization mechanism during training. This approach not only provides valid explanations without additional computational overhead but also enhances model performance and robustness in handling streaming data.

Our proposed SHAP-guided regularization framework bridges this gap by incorporating SHAP-based entropy and stability penalties to encourage sparse and robust feature attributions, making the method applicable to both regression and classification in a unified manner and enhancing generalization while preserving explainability, a crucial factor in real-world decision-making.
In the next section, we formalize our approach, detailing the mathematical formulation, training procedure, and advantages of SHAP-guided regularization.






\section{Methodology}

\subsection{SHAP-Based Regularization for Learning Models}
Our method integrates SHAP values into the model training process by introducing regularization terms based on entropy and stability of the feature attributions. The goal is to improve both the predictive performance and the interpretability of the model by guiding its focus towards the most relevant features while maintaining stable feature importance across similar inputs. This section describes how we incorporate SHAP-guided regularization into the model's loss function.

Given a set of training samples $\{(x_i, y_i)\}$, where $x_i$ represents the feature vector and $y_i$ the target, our objective is to learn a model $f(x_i)$ that minimizes a regularized loss function. For both regression and classification tasks, the total loss function $L_{\text{total}}$ can be defined as:

\begin{equation}
    L_{\text{total}} = L_{\text{task}} + \lambda_1 L_{\text{entropy}} + \lambda_2 L_{\text{stability}}
\end{equation}

where $L_{\text{task}}$ is the standard loss function for the task (e.g., mean squared error for regression or binary cross-entropy for classification). $L_{\text{entropy}}$ is the entropy penalty that encourages sparse and interpretable feature importance distributions. $L_{\text{stability}}$ is the stability penalty that promotes consistency in SHAP attributions across similar samples. $\lambda_1$ and $\lambda_2$ are the regularization hyperparameters that control the influence of the interpretability penalties.

\subsection{Regularization Terms Based on SHAP}

\subsubsection{SHAP Entropy Penalty (Sparsity)}
The entropy penalty $L_{\text{entropy}}$ is designed to sparsify the model's focus on important features. It is calculated as the Shannon entropy of the normalized SHAP values across all features for each prediction:

\begin{equation}
    L_{\text{entropy}} =- \frac{1}{N} \sum_{i=1}^{N}\sum_{j=1}^{M} \hat{p}_{ij} \log (\hat{p}_{ij})
\end{equation}

where: $N$ is the number of samples, $M$ is the number of features, and $\hat{p}_{ij}$ represents the normalized absolute SHAP value for the $j$-th feature in the $i$-th sample. The entropy captures the uncertainty in the feature importance. The penalty encourages models to focus on a small subset of important features, reducing the influence of irrelevant ones. A higher penalty $\lambda_1$ leads to more sparse explanations.

\subsubsection{SHAP Stability Penalty (Consistency)}



The stability regularization term, $L_{\text{stability}}$, is designed to enforce consistency in the model’s explanations by penalizing variations in SHAP values across similar input samples. Specifically, it quantifies how much the attribution of feature importance fluctuates between different but similar data points. Given a dataset of $N$ samples and their associated SHAP values $\phi_{ik}$ for feature $k$ of sample $i$, the stability loss is defined as:

\begin{equation}
    L_{\text{stability}} = \frac{2}{N(N-1)M} \sum_{i=1}^{N} \sum_{\substack{i'=1 \\ i' \neq i}}^{N} \sum_{j=1}^{M} |\phi_{ij} - \phi_{i'j}|,
\end{equation}

where $\phi_{ik}$ and $\phi_{i'k}$ denote the SHAP values for the $j$-th feature of samples $x_i$ and $x_{i'}$ respectively, and $M$ is the number of features. This formulation measures the average pairwise discrepancy in feature attributions across all sample pairs, normalized by the total number of comparisons. By minimizing $L_{\text{stability}}$, the model is encouraged to produce SHAP value distributions that are smooth and consistent across similar instances, thereby enhancing the robustness and reliability of the explanations. The regularization coefficient $\lambda_2$ controls the strength of this penalty; increasing $\lambda_2$ places greater emphasis on producing stable, coherent explanations during model optimization.

Our SHAP-guided regularization method offers several notable advantages. Firstly, by penalizing entropy and enforcing stability, the approach ensures that the model emphasizes the most critical features, leading to sparse and consistent feature attributions. This enhances interpretability, as the model's decisions become more transparent and understandable. Secondly, the incorporation of these regularization terms aids in reducing overfitting. By guiding the model to depend on a smaller, more stable subset of features, it promotes better generalization to unseen data. Thirdly, the framework's flexibility allows its application to both regression and classification tasks, providing a unified approach across different problem domains.




\section{Experiments}

\subsection{Experimental Setup}

We conduct experiments on 10 diverse datasets spanning regression and classification tasks. These datasets vary in size, feature dimensionality, and complexity, ensuring a comprehensive evaluation of our proposed SHAP-guided training approach. Table~\ref{tab:datasets} summarizes the dataset characteristics.

\begin{table}[h]
    \centering
    \caption{Summary of datasets used in experiments, including task type, number of samples, number of features, and target variable.}
    \small
    \begin{tabular}{llccc}
        \toprule
        \textbf{Dataset} & \textbf{Task} & \textbf{Samples} & \textbf{Features} & \textbf{Target Variable} \\
        \midrule
        Diabetes & Regression & 442 & 10 & Disease progression measure \\
        California Housing & Regression & 20,640 & 8 & Median house value \\
        Concrete & Regression & 1,030 & 8 & Concrete compressive strength \\
        Airfoil & Regression & 1,503 & 5 & Sound pressure level \\
        Energy & Regression & 768 & 8 & Heating load \\
        \midrule
        Mushroom & Classification & 8,124 & 22 & Edibility (edible/poisonous) \\
        Banknote Authentication & Classification & 1,372 & 4 & Authenticity (genuine/fake) \\
        Credit Approval & Classification & 690 & 15 & Credit approval status \\
        Breast Cancer & Classification & 569 & 30 & Diagnosis (malignant/benign) \\
        Pima Indians Diabetes & Classification & 768 & 8 & Diabetes status \\
        \bottomrule
    \end{tabular}
    \label{tab:datasets}
\end{table}

\subsubsection{SHAP-Guided Model}
LightGBM \cite{ke2017lightgbm} was selected as the foundational model for implementing SHAP-guided regularization due to several compelling attributes. Its histogram-based algorithm significantly enhances computational efficiency by discretizing continuous feature values into discrete bins, thereby accelerating training processes and reducing memory usage. Additionally, LightGBM's inherent support for TreeSHAP (SHapley Additive exPlanations) facilitates precise estimation of feature importance, making it particularly suitable for interpretability-focused modifications. The model's scalability is another advantage, as it adeptly manages large datasets with extensive feature sets. Furthermore, LightGBM consistently delivers robust performance across both classification and regression tasks. By integrating SHAP-guided regularization into LightGBM, the objective is to harmonize predictive accuracy with enhanced feature interpretability, ensuring that the model not only performs well but also provides transparent insights into its decision-making processes
\paragraph{Model Training Procedure}
Our first exploration of the combined loss function started by applying SHAP-guided regularization within the gradient-boosting framework, specifically using LightGBM for both classification and regression tasks. The training procedure proceeds as follows:

\begin{enumerate}
    \item \textbf{Initialization:} Initialize the LightGBM model with default hyperparameters. Set the regularization hyperparameters $\lambda_1$ and $\lambda_2$ based on experimental settings.
    \item \textbf{Iterative Training:} Train the model using LightGBM’s iterative boosting mechanism. At each iteration $t$, we train a new decision tree and update the model’s parameters.
    \item \textbf{Loss Function Update:} After each boosting iteration, the total loss $L_{\text{total}}$ is computed, which includes the task loss $L_{\text{task}}$, and the regularization terms $L_{\text{entropy}}$ and $L_{\text{stability}}$. The model parameters are then updated to minimize this total loss function.
    \item \textbf{Model Evaluation:} After training, the model is evaluated on a validation set using appropriate metrics (e.g., F1 score and AUC for classification, RMSE for regression).
\end{enumerate}

\paragraph{Hyperparameter Tuning and Optimization}
To fine-tune the performance of the SHAP-guided method, we use cross-validation to select optimal values for $\lambda_1$ and $\lambda_2$. Typically, a grid search or random search is employed to find the combination of hyperparameters that minimizes the combined loss function.

\subsubsection{Evaluation}



 To assess the effectiveness of SHAP-guided regularization, we compare our proposed SHAP-guided LightGBM model against several tree-based baseline machine learning models commonly used for structured data tasks (Decision Tree, Random Forest, LightGBM, XGBoost, and CatBoost).


Our SHAP-guided LightGBM extends the standard LightGBM model by incorporating SHAP-based regularization terms that encourage interpretability and stability in feature attributions.

To evaluate the performance of different models, we utilize the following metrics tailored for regression and classification tasks:

\begin{itemize}
    \item \textbf{Regression:} RMSE (Root Mean Squared Error), $R^2$, SHAP Entropy, Top-k Concentration (Quantifies how concentrated SHAP attributions are among the top-k features), Stability.
    \item \textbf{Classification:} F1-score, AUC (Area Under the Curve), SHAP Entropy, Top-k Concentration, Stability.
\end{itemize}

\subsection{Results}

Tables~\ref{tab:regression_results} and \ref{tab:classification_results} present the aggregated results, evaluating models in terms of standard predictive performance metrics—RMSE and $R^2$ for regression, F1-score and AUC for classification—alongside interpretability-driven metrics, including SHAP Entropy, Top-k Concentration, and Stability. 


\begin{table}[h]
    \centering
      \caption{Aggregated performance results across regression datasets. Lower RMSE and Entropy are better, while higher $R^2$, Top-k Concentration, and Stability indicate better performance and interpretability.}
    \begin{tabular}{lccccc}
        \toprule
        Model & RMSE $\downarrow$ & $R^2$ $\uparrow$ & Entropy $\downarrow$ & Top-k Conc. $\uparrow$ & Stability  $\uparrow$ \\
        \midrule
        Decision Tree & 17.02 & 0.64 & 1.32 & 0.86 & 0.58 \\
        Random Forest & 11.87 & 0.82 & 1.55 & 0.78 & 0.60 \\
        LightGBM & 11.78 & 0.83 & 1.17 & 0.86 & \textbf{0.64} \\
        XGBoost & 12.29 & 0.82 & 1.16 & 0.87 & 0.63 \\
        \textbf{SHAP-guided LightGBM} & \textbf{11.45} & \textbf{0.83} & \textbf{1.12} & \textbf{0.89} & 0.63 \\
        \bottomrule
    \end{tabular}
    \label{tab:regression_results}
\end{table}

\begin{table}[h]
    \centering
      \caption{Aggregated performance results across classification datasets. Lower Entropy is better, while higher F1, AUC, Top-k Concentration, and Stability indicate better performance and interpretability.}
    \small
    \begin{tabular}{lccccc}
        \toprule
        \textbf{Model} & F1 $\uparrow$ & AUC $\uparrow$ & Entropy $\downarrow$ & Top-k Conc. $\uparrow$ & Stability $\uparrow$ \\
        \midrule
        CatBoost            & 0.9194 & 0.9621 & 1.9782 & 0.7898 & \textbf{0.8734} \\
        Decision Tree       & 0.8850 & 0.9197 & \textbf{1.2529} & 0.8722 & 0.8496 \\
        LightGBM           & 0.9141 & 0.9592 & 1.8261 & 0.8551 & 0.8647 \\
        Random Forest       & 0.9163 & 0.9604 & 2.1783 & 0.7382 & 0.8699 \\
        XGBoost             & 0.9171 & 0.9615 & 1.8904 & 0.7993 & 0.8716 \\
         \textbf{SHAP-guided LightGBM }&\textbf{ 0.9207} &\textbf{0.9641} & 1.6542 & \textbf{0.8905} & 0.8604 \\
        \bottomrule
    \end{tabular}
    \label{tab:classification_results}
\end{table}

For regression tasks, the SHAP-guided LightGBM maintains competitive predictive performance while improving interpretability. The model achieves an RMSE of 11.45, which is comparable to standard LightGBM (11.78) and outperforms other baselines. Similarly, the $R^2$ score remains at 0.83, confirming that the model retains its ability to explain variance in the data. In terms of interpretability, SHAP Entropy is reduced to 1.12, indicating that feature importance is more concentrated and less dispersed compared to standard LightGBM (1.17) and Random Forest (1.55). This suggests that SHAP-guided regularization encourages a more structured attribution pattern, enhancing transparency in feature importance. Furthermore, Top-k Concentration improves to 0.89, surpassing the standard LightGBM (0.86) and XGBoost (0.87), meaning that the model places greater emphasis on the most relevant features. Stability remains at 0.63, aligning closely with baseline models, demonstrating that the regularization does not introduce fluctuations in feature attributions.

For classification tasks, similar trends are observed. The SHAP-guided LightGBM achieves an F1-score of 0.9207 and an AUC of 0.9641, both slightly surpassing the standard LightGBM (0.9141 F1-score, 0.9592 AUC). This indicates that the introduction of SHAP-based regularization does not degrade predictive performance. More importantly, SHAP Entropy is reduced to 1.6542, compared to 1.8261 for LightGBM and 2.1783 for Random Forest, highlighting a more refined and focused attribution distribution. Top-k Concentration is the highest among all models (0.8905), confirming that the model consistently assigns importance to a small subset of critical features, which enhances interpretability. Stability remains competitive at 0.8604, slightly lower than LightGBM (0.8647) but higher than other baselines, ensuring robustness in feature attributions.

\begin{figure}[htbp]
    \centering
    \begin{subfigure}[b]{0.8\textwidth}
        \centering
        \includegraphics[width=\textwidth]{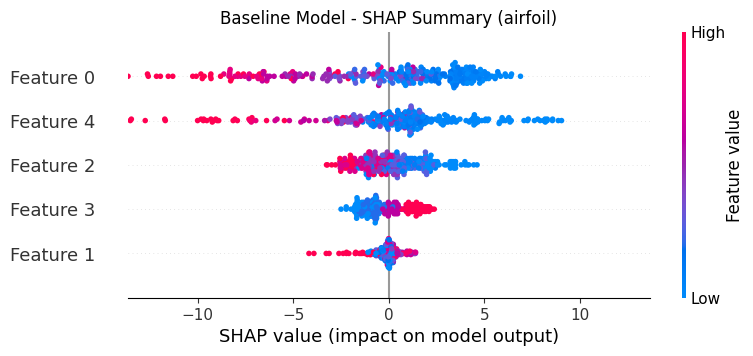}
        \caption{Baseline lightGBM}
        \label{fig:sub1}
    \end{subfigure}
    \hfill
    \begin{subfigure}[b]{0.8\textwidth}
        \centering
        \includegraphics[width=\textwidth]{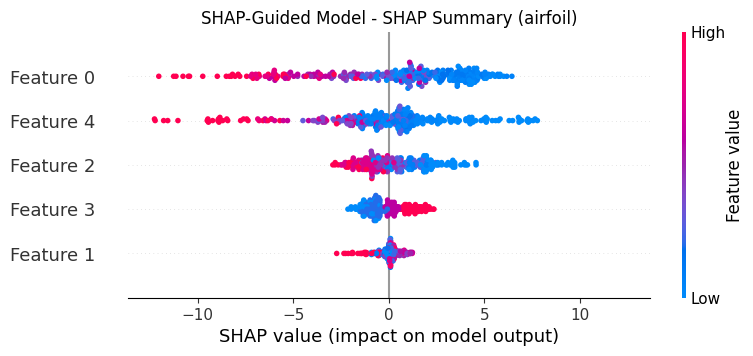}
        \caption{SHAP-guided lightGBM}
        \label{fig:sub2}
    \end{subfigure}
    \caption{SHAP Diagram on the airfoil regression dataset.}
    \label{fig:main}
\end{figure}

\begin{figure}
        \centering
        \includegraphics[width=0.5\textwidth]{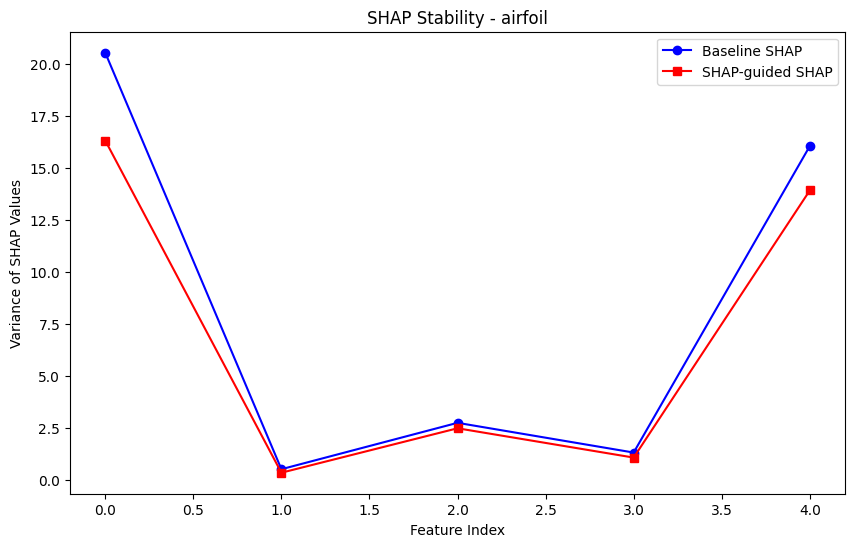}
        \caption{SHAP values mean variance comparison on the airfoil regression dataset.}
        \label{fig:main2}
\end{figure}

Figure \ref{fig:main} shows an illustration of SHAP diagram using standard lightGBM (Baseline Model) and the SHAP-guided LightGBM for the airfoil regression dataset. It is clear that the SHAP regularization promotes stability by compromising similar feature importance to similar samples (more condensed regions in Figure \ref{fig:sub2}). This is confirmed further by Figure \ref{fig:main2}, which shows lower variance of SHAP values across different features using the guided-SHAP model for the same dataset.

Overall, these results demonstrate that SHAP-guided regularization effectively enhances interpretability without compromising predictive accuracy. The method successfully reduces SHAP Entropy, leading to sparser and more meaningful feature attributions, while increasing Top-k Concentration, ensuring the model prioritizes the most relevant features. Stability remains comparable to non-regularized models, confirming that the proposed method does not introduce instability in feature attributions. These findings indicate that SHAP-guided learning can serve as a powerful tool for balancing interpretability and predictive performance in tree-based models.





\section{Conclusion}
We introduced a first exploration of SHAP-guided loss training. First experiments on LightGBM showed that SHAP-based regularization promotes interpretable and stable feature attributions while maintaining strong predictive performance.

SHAP regularization requires further detailed exploration as it seems to be a potential tool for:

\begin{itemize}
    \item Improving SHAP-based interpretability metrics without degrading accuracy.
    \item Enhancing feature attribution stability across datasets.
    \item Providing a novel approach to balancing predictive performance with interpretability in ML models.
\end{itemize}

These insights demonstrate that SHAP-guided learning is a promising direction for explainable machine learning.

\bibliography{sample-ceur}

\begin{thebibliography}{17}
\expandafter\ifx\csname natexlab\endcsname\relax\def\natexlab#1{#1}\fi
\providecommand{\url}[1]{\texttt{#1}}
\providecommand{\href}[2]{#2}
\providecommand{\path}[1]{#1}
\providecommand{\DOIprefix}{doi:}
\providecommand{\ArXivprefix}{arXiv:}
\providecommand{\URLprefix}{URL: }
\providecommand{\Pubmedprefix}{pmid:}
\providecommand{\doi}[1]{\href{http://dx.doi.org/#1}{\path{#1}}}
\providecommand{\Pubmed}[1]{\href{pmid:#1}{\path{#1}}}
\providecommand{\bibinfo}[2]{#2}
\ifx\xfnm\relax \def\xfnm[#1]{\unskip,\space#1}\fi
\bibitem[{Das and Rad(2020)}]{das2020opportunities}
\bibinfo{author}{A.~Das}, \bibinfo{author}{P.~Rad},
\newblock \bibinfo{title}{Opportunities and challenges in explainable artificial intelligence (xai): A survey},
\newblock \bibinfo{journal}{arXiv preprint arXiv:2006.11371}  (\bibinfo{year}{2020}).
\bibitem[{Molnar(2020)}]{molnar2020interpretable}
\bibinfo{author}{C.~Molnar}, \bibinfo{title}{Interpretable machine learning}, \bibinfo{publisher}{Lulu. com}, \bibinfo{year}{2020}.
\bibitem[{Belle and Papantonis(2021)}]{belle2021principles}
\bibinfo{author}{V.~Belle}, \bibinfo{author}{I.~Papantonis},
\newblock \bibinfo{title}{Principles and practice of explainable machine learning},
\newblock \bibinfo{journal}{Frontiers in big Data} \bibinfo{volume}{4} (\bibinfo{year}{2021}) \bibinfo{pages}{688969}.
\bibitem[{Li(2022)}]{li2022extracting}
\bibinfo{author}{Z.~Li},
\newblock \bibinfo{title}{Extracting spatial effects from machine learning model using local interpretation method: An example of shap and xgboost},
\newblock \bibinfo{journal}{Computers, Environment and Urban Systems} \bibinfo{volume}{96} (\bibinfo{year}{2022}) \bibinfo{pages}{101845}.
\bibitem[{Altmann et~al.(2010)Altmann, Tolo{\c{s}}i, Sander, and Lengauer}]{altmann2010permutation}
\bibinfo{author}{A.~Altmann}, \bibinfo{author}{L.~Tolo{\c{s}}i}, \bibinfo{author}{O.~Sander}, \bibinfo{author}{T.~Lengauer},
\newblock \bibinfo{title}{Permutation importance: a corrected feature importance measure},
\newblock \bibinfo{journal}{Bioinformatics} \bibinfo{volume}{26} (\bibinfo{year}{2010}) \bibinfo{pages}{1340--1347}.
\bibitem[{Menze et~al.(2009)Menze, Kelm, Masuch, Himmelreich, Bachert, Petrich, and Hamprecht}]{menze2009comparison}
\bibinfo{author}{B.~H. Menze}, \bibinfo{author}{B.~M. Kelm}, \bibinfo{author}{R.~Masuch}, \bibinfo{author}{U.~Himmelreich}, \bibinfo{author}{P.~Bachert}, \bibinfo{author}{W.~Petrich}, \bibinfo{author}{F.~A. Hamprecht},
\newblock \bibinfo{title}{A comparison of random forest and its gini importance with standard chemometric methods for the feature selection and classification of spectral data},
\newblock \bibinfo{journal}{BMC bioinformatics} \bibinfo{volume}{10} (\bibinfo{year}{2009}) \bibinfo{pages}{1--16}.
\bibitem[{Antwarg et~al.(2021)Antwarg, Miller, Shapira, and Rokach}]{antwarg2021explaining}
\bibinfo{author}{L.~Antwarg}, \bibinfo{author}{R.~M. Miller}, \bibinfo{author}{B.~Shapira}, \bibinfo{author}{L.~Rokach},
\newblock \bibinfo{title}{Explaining anomalies detected by autoencoders using shapley additive explanations},
\newblock \bibinfo{journal}{Expert systems with applications} \bibinfo{volume}{186} (\bibinfo{year}{2021}) \bibinfo{pages}{115736}.
\bibitem[{Schmidt et~al.(2009)Schmidt, Fung, and Rosales}]{schmidt2009optimization}
\bibinfo{author}{M.~Schmidt}, \bibinfo{author}{G.~Fung}, \bibinfo{author}{R.~Rosales},
\newblock \bibinfo{title}{Optimization methods for l1-regularization},
\newblock \bibinfo{journal}{University of British Columbia, Technical Report TR-2009-19}  (\bibinfo{year}{2009}).
\bibitem[{Cortes et~al.(2012)Cortes, Mohri, and Rostamizadeh}]{cortes2012l2}
\bibinfo{author}{C.~Cortes}, \bibinfo{author}{M.~Mohri}, \bibinfo{author}{A.~Rostamizadeh},
\newblock \bibinfo{title}{L2 regularization for learning kernels},
\newblock \bibinfo{journal}{arXiv preprint arXiv:1205.2653}  (\bibinfo{year}{2012}).
\bibitem[{Chen et~al.(2011)Chen, Miao, Wang, and Wu}]{chen2011rough}
\bibinfo{author}{Y.~Chen}, \bibinfo{author}{D.~Miao}, \bibinfo{author}{R.~Wang}, \bibinfo{author}{K.~Wu},
\newblock \bibinfo{title}{A rough set approach to feature selection based on power set tree},
\newblock \bibinfo{journal}{Knowledge-Based Systems} \bibinfo{volume}{24} (\bibinfo{year}{2011}) \bibinfo{pages}{275--281}.
\bibitem[{Brauwers and Frasincar(2021)}]{brauwers2021general}
\bibinfo{author}{G.~Brauwers}, \bibinfo{author}{F.~Frasincar},
\newblock \bibinfo{title}{A general survey on attention mechanisms in deep learning},
\newblock \bibinfo{journal}{IEEE Transactions on Knowledge and Data Engineering} \bibinfo{volume}{35} (\bibinfo{year}{2021}) \bibinfo{pages}{3279--3298}.
\bibitem[{Alvarez~Melis and Jaakkola(2018)}]{alvarez2018towards}
\bibinfo{author}{D.~Alvarez~Melis}, \bibinfo{author}{T.~Jaakkola},
\newblock \bibinfo{title}{Towards robust interpretability with self-explaining neural networks},
\newblock \bibinfo{journal}{Advances in neural information processing systems} \bibinfo{volume}{31} (\bibinfo{year}{2018}).
\bibitem[{Lundberg and Lee(2017)}]{lundberg2017unified}
\bibinfo{author}{S.~M. Lundberg}, \bibinfo{author}{S.-I. Lee},
\newblock \bibinfo{title}{A unified approach to interpreting model predictions},
\newblock \bibinfo{journal}{Advances in neural information processing systems} \bibinfo{volume}{30} (\bibinfo{year}{2017}).
\bibitem[{Wang et~al.(2021)Wang, Wang, and Inouye}]{wang2021shapley}
\bibinfo{author}{R.~Wang}, \bibinfo{author}{X.~Wang}, \bibinfo{author}{D.~I. Inouye},
\newblock \bibinfo{title}{Shapley explanation networks},
\newblock \bibinfo{journal}{arXiv preprint arXiv:2104.02297}  (\bibinfo{year}{2021}).
\bibitem[{Sevillano-Garc{\'\i}a et~al.(2024)Sevillano-Garc{\'\i}a, Luengo, and Herrera}]{sevillano2024x}
\bibinfo{author}{I.~Sevillano-Garc{\'\i}a}, \bibinfo{author}{J.~Luengo}, \bibinfo{author}{F.~Herrera},
\newblock \bibinfo{title}{X-shield: Regularization for explainable artificial intelligence},
\newblock \bibinfo{journal}{arXiv preprint arXiv:2404.02611}  (\bibinfo{year}{2024}).
\bibitem[{Cheng et~al.(2023)Cheng, Qu, and Lee}]{cheng2023shapnn}
\bibinfo{author}{Q.~Cheng}, \bibinfo{author}{S.~Qu}, \bibinfo{author}{J.~Lee},
\newblock \bibinfo{title}{Shapnn: Shapley value regularized tabular neural network},
\newblock \bibinfo{journal}{arXiv preprint arXiv:2309.08799}  (\bibinfo{year}{2023}).
\bibitem[{Ke et~al.(2017)Ke, Meng, Finley, Wang, Chen, Ma, Ye, and Liu}]{ke2017lightgbm}
\bibinfo{author}{G.~Ke}, \bibinfo{author}{Q.~Meng}, \bibinfo{author}{T.~Finley}, \bibinfo{author}{T.~Wang}, \bibinfo{author}{W.~Chen}, \bibinfo{author}{W.~Ma}, \bibinfo{author}{Q.~Ye}, \bibinfo{author}{T.-Y. Liu},
\newblock \bibinfo{title}{Lightgbm: A highly efficient gradient boosting decision tree},
\newblock \bibinfo{journal}{Advances in neural information processing systems} \bibinfo{volume}{30} (\bibinfo{year}{2017}).

\end{thebibliography}




\end{document}